# Arabic Character Segmentation Using Projection-Based Approach with Profile's Amplitude Filter

Mahmoud A. A. Mousa
Dept. of Computer and Systems Engineering,
Zagazig University, Zagazig, Egypt
mamosa@zu.edu.eg

Mohammed S. Sayed and Mahmoud I. Abdalla
Dept. of Electronics and Communications Engineering,
Zagazig University, Zagazig, Egypt
msayed@zu.edu.eg, mabdalla@zu.edu.eg

*Abstract*—Arabic is one of the languages that present special challenges to Optical character recognition (OCR). The main challenge in Arabic is that it is mostly cursive. Therefore, a segmentation process must be carried out to determine the character's start and end. This step is essential for character recognition. This paper presents Arabic character segmentation algorithm. The proposed algorithm uses the projection-based approach concepts to separate lines, words, and characters. This is done using profile's amplitude filter and simple edge tool to find characters separations. Our algorithm shows promising performance when applied on different printed documents with different Arabic fonts.

I. INTRODUCTION

Optical character recognition (OCR) is an application for image recognition that studies automatic reading. This is done by taking an image of text written in a specific language to be understood by the computer and get the final computer representation for this text. OCR techniques may vary according to the language which will be used, its nature and the application in which this technique is applied [1]. The ultimate goal of OCR is to imitate the human ability to read at a much faster rate by associating symbolic identities with images of characters.

Arabic is one of the languages that present special challenges to OCR. The main challenge in Arabic is that it is mostly cursive. Arabic is written by connecting characters together to produce words or parts of words as shown in Fig. 1. Arabic text is written from right to left. Arabic language has 28 basic characters, of which 16 have from one to three dots.

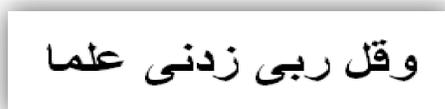

Figure 1. The characters connectivity of Arabic text.

Arabic characters have many shapes and depend mainly on their position in the word. For example, the character "noon" is written in the form of "نـ" at the start, "ـنـ" at the middle, and "ـن" at the end of a word but the separated form of this character is "ن". The shape and the size of Arabic characters vary with respect to their position in the word and this is a great challenge in Arabic text [1].

Because of the different nature Arabic text fonts, characters may overlap vertically to produce certain compounds of characters at certain positions of the Arabic word segments such as "محـ , حمـ , نجـ" which can be represented by single atomic graphemes called ligatures. Traditional Arabic font for example contains around 220 graphemes, and another common less involved font (with fewer ligatures) like Simplified Arabic contains around 151 graphemes [1, 17, 18].

Some Arabic characters have single dot such as "ن , ج , ب" and another characters have double dots such as "تـ , يـ" and other characters have triple dots such as "ثـ , شـ". The doted characters exhibit a big problem while being processed.

This paper presents Arabic character segmentation algorithm. The proposed algorithm uses the projection-based approach concepts to separate lines, word, and characters using profile's amplitude filter and simple edge tool. The rest of the paper is organized as follows: Section 2 reviews different segmentation techniques. Section 3 presents the proposed algorithm. Section 4 demonstrates the results and performance analysis. Section 5 concludes this paper.

II. SEGMENTATION TECHNIQUES

In this part, methods of how to convert the image that contains Arabic text into character images are discussed. This is done using three segmentation stages: line segmentation, word segmentation, and character segmentation.

*A. Line segmentation approaches:*

*Projection-based approach;* in which pixels of image are being summed along the horizontal axis for each y value and this is referred as a horizontal projection [2-5, 10,14-18 ] or along the vertical axis for each x value on the segmented line image and this is called vertical projection [2, 3, 12, 13, 14-18].

*Smearing approach;* in which consecutive black pixels along the horizontal direction are smeared. The distance between the white space is calculated. If the distance lies within a predefined threshold, it is filled with black pixels. The text lines are bounded with connected shapes of black pixels [6, 10, 11].

*Grouping approach;* in which text lines are iteratively constructed by grouping neighboring connected components based on certain perceptual criteria such as similarity, continuity and proximity [7, 10].

*Hough-based approach* in which the Hough transform is used for locating straight lines in text images [8-10].

### B. Word and character segmentation approaches:

There are four main approaches that deal with connected characters in an Arabic word [1].
  a. Assuming that the input is already segmented into characters (i.e., no character segmentation will be needed).
  b. Segmenting input words into primitives smaller than a character then collecting each group of primitives into character while being recognized.
  c. Segmenting words into characters. This is the most difficult approach in cursive nature languages.
  d. Recognizing input words, as a whole, with no segmentation.

### C. Histogram-based Algorithms

Several algorithms use histogram-based techniques for OCR in different languages. None of these algorithms solved all the problems associated with OCR in Arabic language. In Telugu script, the text document image may contain overlapped lines and characters and no cursive nature between characters [15,21]. Devnagari is used for writing Hindi, Marathi, Sanskrit and Nepali languages. Characters may be separate or connected with a horizontal line at the upper part, known as Shirorekha and no overlapping challenge considered [16]. Only line and connected parts segmentation is performed on Arabic language and the main difficult in Arabic language, which is to separate characters and this, is not considered in [20].

### III. THE PROPOSED ALGORITHM

This paper concerned with segmentation procedure that accepts an Arabic text image and outputs separated characters. It has three stages. The first one is responsible for detecting and separating lines in the text. The segmented lines are then passed throw the second stage which is designed to get words from text lines. Finally, the third stage takes those words and produces the character representation of each word. The proposed algorithm considers the cursive and the partially overlapping natures between characters .A pre-stage is used for detecting and correcting the skew for the scanned text image. The algorithm in [19] is used to perform the skew angle correction.

### A. Line segmentation:

Line segmentation is done using Image Axis Profile method that calculates the horizontal axis profile for the binarized text image [10]. The horizontal axis profile matrix $I_j$ is calculated by summing pixels values P(i,j) along the X-axis for each y value as shown in (1).

$$I_j = \sum_{i=0}^{i=n}(255 - P(i,j)) \quad (1)$$

where i, j are X and Y - axis indices respectively, n = X-resolution.

This profile has information about the text lines that are indicated by the regions with the black intensities as in Fig. 2. On the other hand, the blank lines appear as a drop in the black intensities. The text lines can be extracted by comparing the profile with a pre-defined threshold and this can be achieved by performing the algorithm described in Fig. 3.

Fig. 3 shows the line segmentation algorithm that accepts an image written in Arabic and extracts its lines. This is done using the horizontal axis profile on two stages. The first one is to locate each connected group of dark regions in the profile. The other one is to decide which dark region(s) can be considered as a separate line.

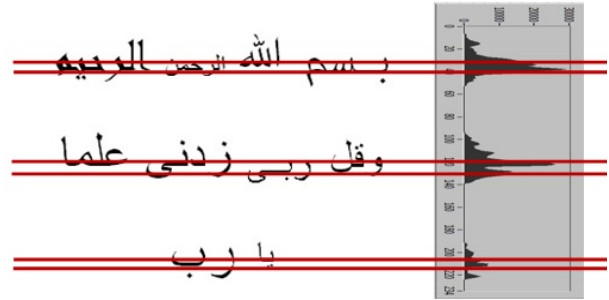

Figure 2.  Horizontal axis profile for a sample image with different fonts and different word size.

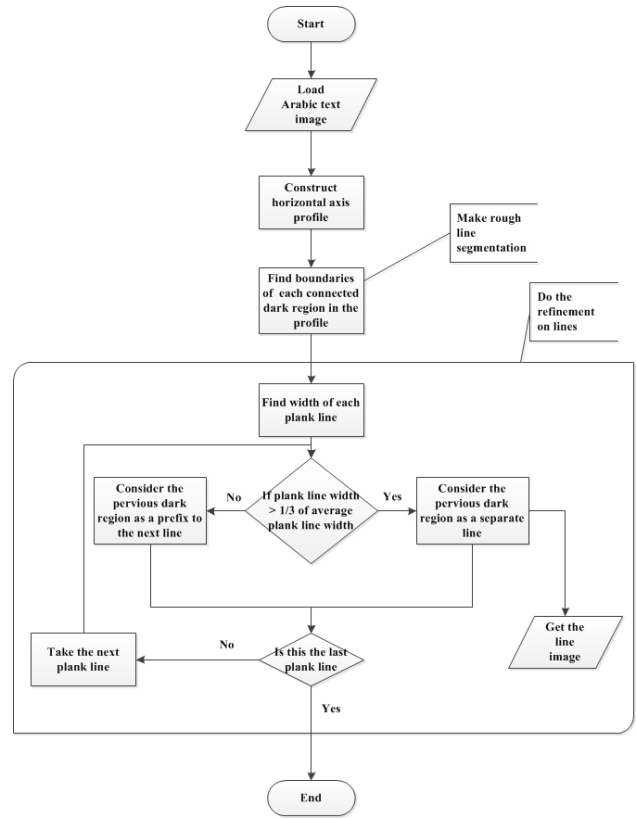

Figure 3.  Line segmentation algorithm.

     123

## B. Word segmentation:

Fig. 4 shows an Arabic text line image and its corresponding vertical profile. The text line is segmented, beginning from the left side to the right, into connected parts. These connected parts are clustered to the corresponding word. The algorithm in explained in Fig. 5. Each word is an input image to the character segmentation stage.

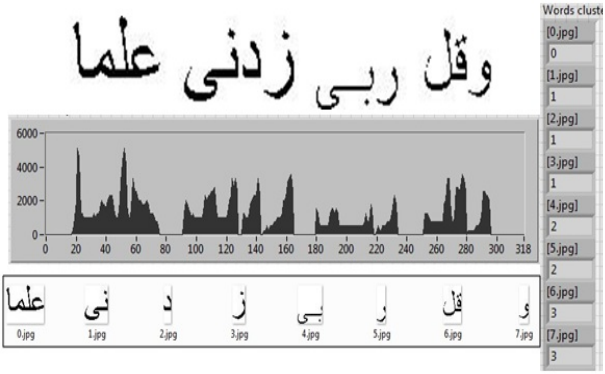

Figure 4. A test sample for the word segmentation algorithm

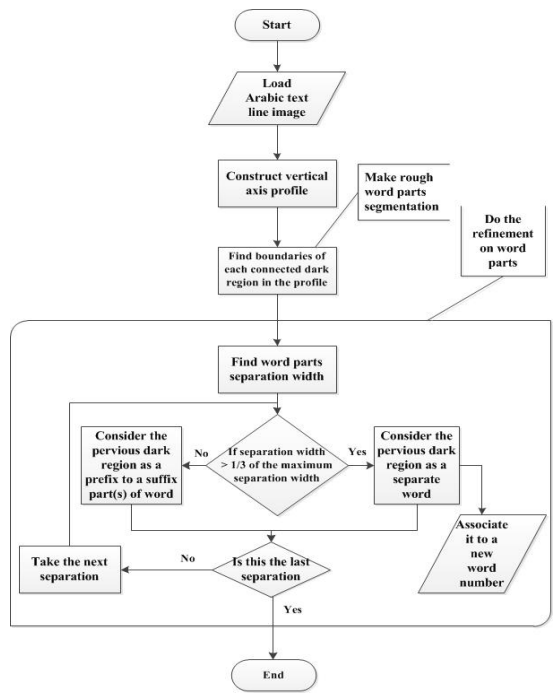

Figure 5. Word parts segmentation algorithm

## C. Character segmentation:

Character segmentation process is the most important one in the OCR system, because character will be then entered to the recognition stage so it should be correctly separated with no error to be recognized correctly. As shown in Fig. 6, the text word/sub-word image is entered to the character segmentation stage which calculates the base-line coordinates for each text line image as shown in Fig. 2, which is in red color. The base-line is the space around the maximum value in the horizontal profile [12, 15, 17].

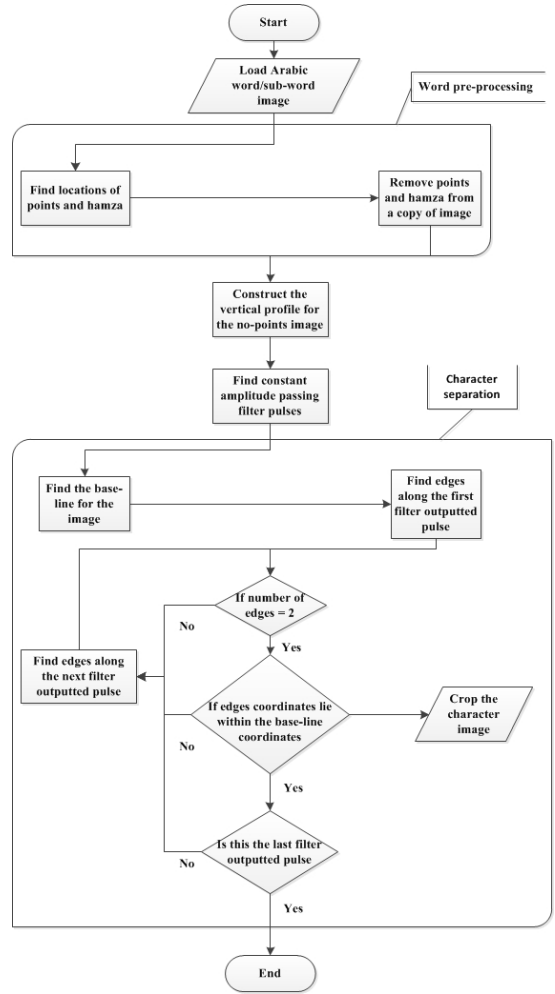

Figure 6. The character segmentation algorithm

The proposed algorithm mainly depends on the vertical profile and its dc components that correspond to connections between two characters. So, the existence of Hamza of Alif character such as أ , إ , لا , points of the characters such as ن , تـ , ثـ , شـ and all other characters that have points above or under the character shape may produce errors in our technique, so another no-points and Hamza image is to be constructed from the previous word image . The method of locating points and Hamza is to locate connected group of dark pixels. The points and Hamza are small connected areas above or under the baseline. So, they can be erased successfully as shown in Fig. 7.

By constructing the vertical profile for the no points word image. The separation between two characters is considered as constant amplitude in the profile. A constant amplitude (low variation) passing filter is designed so that only low variations in the profile will be passed. The filter's output pulses are a locus of the characters connections. This locus takes the shape of separated sequential train of pulses as shown in Fig. 7.

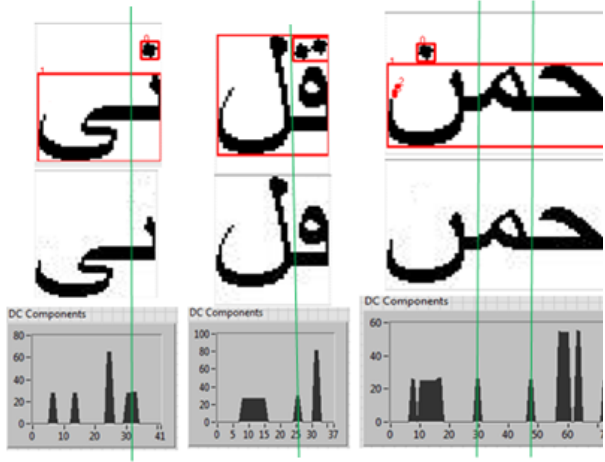

Figure 7.  A test sample for the character segmentation algorithm

Each pulse in the locus shape will be checked to determine whether it is a correct characters connection or not. This is done by using the simple edge tool which finds step edges along an array of pixel coordinates. This tool returns the coordinates of all the edges found. If there is only two edges and the coordinates of these edges lie within the base-line coordinates, then this pulse is considered as a character connector as shown in Fig. 7.

To calculate the base-line coordinates, the vertical index corresponding to the maximum peek in the horizontal axis profile $V\text{-}ind_{max}$ is calculated for each text line generated from the line segmentation stage as shown in (3).

$$V\text{-}ind_{max} = IndexOf(\max(I)) \qquad (3)$$

This value is bounded by two lines which are the base line coordinates as shown in Fig. 2.

An image cutter tool is used on the original text image to extract sub-images which correspond to each pulse in the filter response. Each sub-image is extracted vertically starting from the first index to the last index of each pulse. A horizontal axis profile is calculated for each sub-image. If the output of the horizontal axis profile takes the shape of one pulse with constant amplitude and the vertical index $V\text{-}ind_{max}$ lies inside this pulse, then the pulse coordinates are considered as the base line coordinates.

Fig. 8 shows a word, its filter output, and the horizontal axis profile for each sub-image corresponds to each pulse in the filter output. The vertical index for the second line in Fig. 2 is $V\text{-}ind_{max} = 29$. Hence, the first vertical pulse in Fig. 8 is refused because it gives a horizontal profile from 36 to 41, which is away from $V\text{-}ind_{max}$. The second vertical pulse is accepted because the vertical index $V\text{-}ind_{max}$ lies inside its horizontal profile (i.e. from 28 to 32). The third vertical pulse is refused as it gives more than one pulse in its horizontal axis profile.

Fig. 7 shows the original test images, the no point images, and output of the low variation passing filter whose input is the no-point vertical profile for the word. For the pervious algorithm, only pulses pass are those which marked with a green line. Character separation operation is done at the green color lines for the original image with points and Hamza.

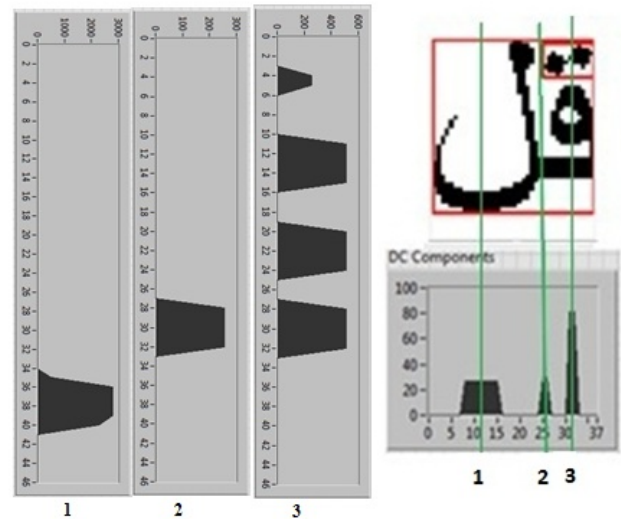

Figure 8.  A word with its filter output and the horizontal axis profile for each sub-image corresponds to each pulse

The character (ر) may overlap in many fonts with the remaining part of the word. This represents a challenge when separating the word into connected parts. The proposed algorithm locate these connected parts as well as points and Hamza overcoming the problem of separating partially overlapped characters in [15] as shown in Fig. 9.

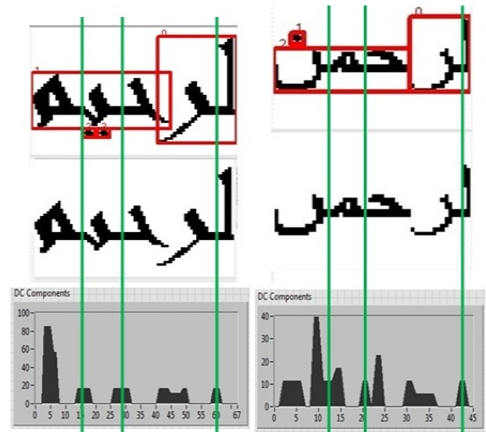

Figure 9.  Samples with different font and size

IV.  RESULTS AND PERFORMANCE EVALUATION

Test operation is done on a document (264 pages, 6,286 lines, 50,931 words, 224,781 characters with no spaces, 275,136 characters with spaces). The line segmentation algorithm achieved a very large correction ratio that reaches 99.9%. One error is found in the test when dealing with a group of text lines bounded by a text box. All these lines and the text box are treated as a single line and this happened because of the continuity of the text box.

The word segmentation algorithm achieved a correction ratio of approximately 99.9%. One error appeared when dealing with marks such as exclamation, question, comma, and semicolon. When writing those marks directly after the word with no separation space (e.g. ماذا؟ , ! ,متى), the algorithm gives an indication that those marks belong to the same pervious word cluster number as the space between them is in sufficient to decide that the mark is another word.

The character segmentation algorithm achieved also an acceptable correction ratio that reaches 98%. This algorithm deals with the compounded characters as a separate character (e.g. محـ , نجـ). These compounded characters will be dealt with in the recognition stage. This makes the alphabetic Arabic language equal to 28 basic characters plus 220 compounded characters.

## V. CONCLUSION

This paper presents Arabic character segmentation algorithm. The proposed algorithm uses the projection-based approach concepts to separate lines, words, and characters. The lines produced from the line segmentation stage are entered to the next one which segments them to connected parts. These connected parts may be separate characters or a number of connected characters. Those connected characters are entered then to the character segmentation stage that is responsible for separating these connected characters. The proposed algorithm achieved a promising success rate ratio for line, word and character segmentation.